\documentclass[10pt,twocolumn,letterpaper]{article}

\usepackage{iccv}
\usepackage{times}
\usepackage{epsfig}
\usepackage{graphicx}
\usepackage{amsmath}
\usepackage{amssymb}
\usepackage{booktabs}
\usepackage{capt-of}
\usepackage{enumitem}

\usepackage{array}
\usepackage{color, xcolor,colortbl}
\usepackage{makecell}
\usepackage{multirow}
\usepackage[square,comma,numbers,sort&compress]{natbib}
\usepackage[normalem]{ulem}
\usepackage{authblk}

\usepackage[pagebackref=true,breaklinks=true,letterpaper=true,colorlinks,bookmarks=false]{hyperref}

\iccvfinalcopy 

\newcommand{\minival}{minival\raisebox{0.2ex}{$\ast$}}

\def\iccvPaperID{1498} 

\ificcvfinal\fi
\begin{document}

\title{Revisiting Unreasonable Effectiveness of Data in Deep Learning Era}

\author[1]{Chen Sun}
\author[1,2]{Abhinav Shrivastava}
\author[1]{Saurabh Singh}
\author[1,2]{Abhinav Gupta}

\affil[1]{Google Research}
\affil[2]{Carnegie Mellon University}

\maketitle

\begin{abstract}
The success of deep learning in vision can be attributed to: (a) models with high capacity; (b) increased computational power; and (c) availability of large-scale labeled data. Since 2012, there have been significant advances in representation capabilities of the models and computational capabilities of GPUs. But the size of the biggest dataset has surprisingly remained constant.
What will happen if we increase the dataset size by $10\times$ or $100\times$? This paper takes a step towards clearing the clouds of mystery surrounding the relationship between `enormous data' and visual deep learning. 
By exploiting the JFT-300M dataset which has more than 375M noisy labels for 300M images, we investigate how the performance of current vision tasks would change if this data was used for representation learning.
Our paper delivers some surprising (and some expected) findings. First, we find that the performance on vision tasks increases logarithmically based on volume of training data size. Second, we show that representation learning (or pre-training) still holds a lot of promise. One can improve performance on many vision tasks by just training a better base model. Finally, as expected, we present new state-of-the-art results for different vision tasks including image classification, object detection, semantic segmentation and human pose estimation. Our sincere hope is that this inspires vision community to 
not undervalue the data and develop collective efforts in building larger datasets.

\vspace{-0.1in}
\end{abstract}

\section{Introduction}

\begin{figure}[h]
\center
\includegraphics[width=0.9\columnwidth]{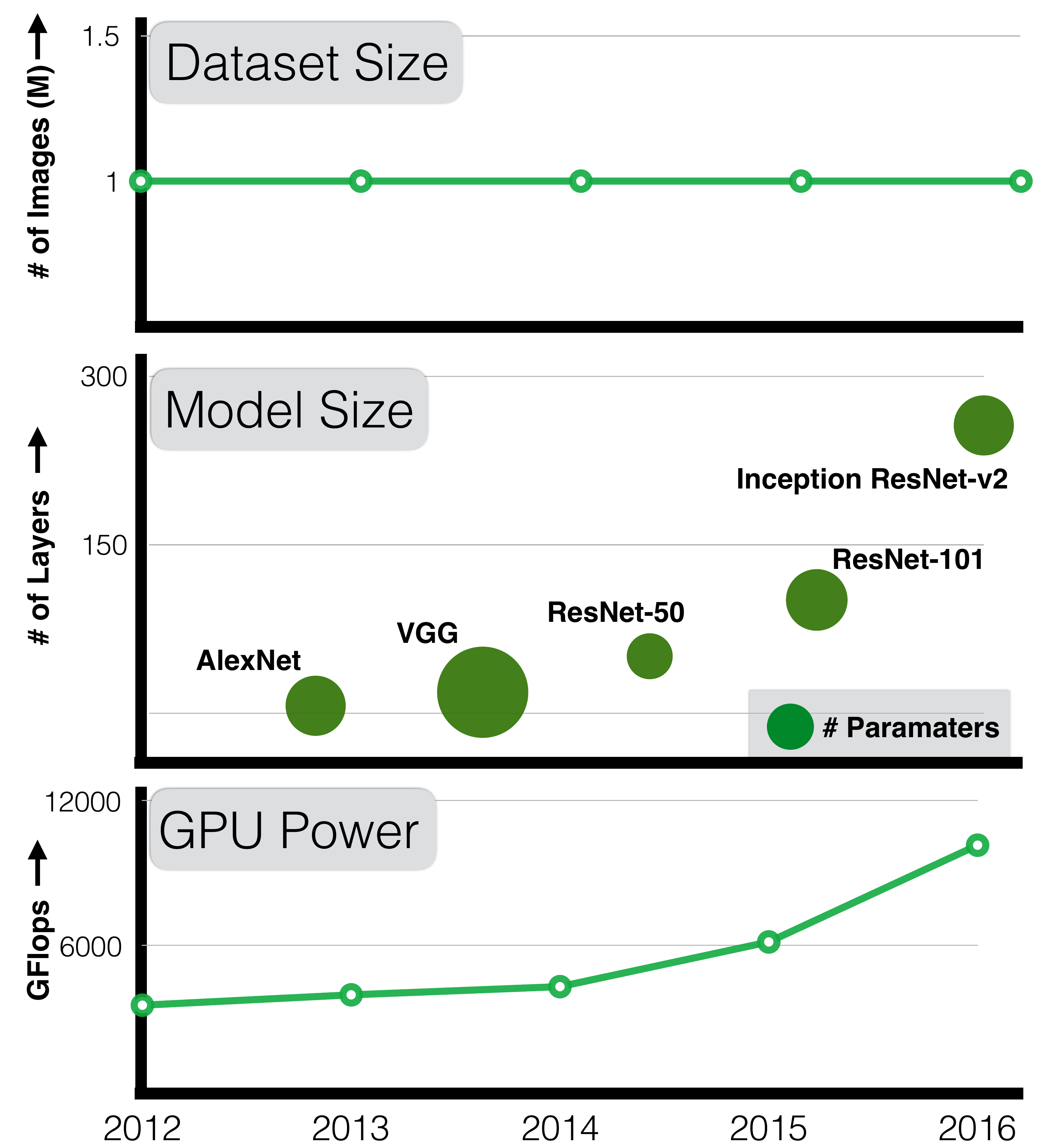}
\caption{The Curious Case of Vision Datasets: While GPU computation power and model sizes have continued to increase over the last five years, size of the largest training dataset has surprisingly remained constant. Why is that? What would have happened if we have used our resources to increase dataset size as well? This paper provides a sneak-peek into what could be if the dataset sizes are increased dramatically.}
\vspace{-0.15in}
\label{fig:teasers}
\end{figure}

There is unanimous agreement that the current ConvNet revolution is a product of big labeled datasets (specifically, 1M labeled images from ImageNet~\cite{ImageNet}) and large computational power (thanks to GPUs). Every year we get further increase in computational power (a newer and faster GPU) but our datasets have not been so fortunate. ImageNet, a dataset of 1M labeled images based on 1000 categories, was used to train AlexNet~\cite{AlexNet} more than five years ago. Curiously, while both GPUs and model capacity
have continued to grow, datasets to train these models have remained stagnant. Even a 101-layer ResNet with significantly more capacity and depth is still trained with 1M images from ImageNet circa 2011.
Why is that? Have we once again belittled the importance of data in front of deeper models and computational power? What will happen if we scale up the amount of training data $10\times$ or $100\times$, will the performance double?

This paper takes the first steps towards clearing the clouds of mystery surrounding the relationship between `enormous data' and deep learning. 
We exploit the already existing JFT-image dataset, first introduced by Hinton \etal~\cite{Hinton} and expanded by~\cite{Chollet}. The JFT dataset has more than 300M images that are labeled with 18291 categories. The annotations have been automatically obtained and, therefore, are noisy and not exhaustive. These annotations have been cleaned using complex algorithms to increase the precision of labels; however there is still approximately 20\% error in precision. We will use this data to investigate the nature of relationship between amount of data and performance on vision tasks. Specifically, we will look into the power of data for visual representation learning (pre-training). 
We evaluate our learned representation on a variety of vision tasks: image classification, object detection, semantic segmentation and human pose estimation. Our experiments yield some surprising (and some expected) findings:

\begin{itemize}
    \itemsep0em 
    \item {\bf Better Representation Learning Helps!} Our first observation is that large-scale data helps in representation learning as evidenced by improvement in performance on each and every vision task we study.
    
    This suggests that collection of a larger-scale dataset to study visual pretraining may greatly benefit the field. Our findings also suggest a bright future for unsupervised or self-supervised~\cite{DBLP:journals/corr/WangG15a, doersch2015unsupervised} representation learning approaches. It seems the scale of data can overpower noise in the label space.
    \item {\bf Performance increases logarithmically based on volume of training data.} We find there is a logarithmic relationship between performance on vision tasks and the amount of training data used for representation learning. Note that previous papers on large-scale learning~\cite{JoulinWeakly} have shown diminishing returns even on log-scale.
    
    \item {\bf Capacity is Crucial:} We also observe that to fully exploit 300M images, one needs higher capacity models. For example, in case of ResNet-50 the gain on COCO object detection is much smaller (1.87\%) compared to (3\%) when using ResNet-152.
    
    \item {\bf Training with Long-tail:} Our data has quite a long tail and yet the representation learning seems to work. This long-tail does not seem to adversely affect the stochastic training of ConvNets (training still converges).
    
    \item {\bf New state of the art results:} Finally, our paper presents new state-of-the-art results on several benchmarks using the models learned from JFT-300M. For example, a single model (without any bells and whistles) can now achieve 37.4 AP as compared to 34.3 AP on the COCO detection benchmark.
\end{itemize}

\section{Related Work}
Ever since the seminal work by Krizhevsky~\etal~\cite{AlexNet} showcased the power of Convolutional Neural Networks (ConvNets) on large-scale image recognition task, a lot of work has been done to make them more accurate. A common approach is to increase the complexity of these networks by increasing the width or depth of these networks. For example, Simonyan and Zisserman~\cite{Simonyan14VGG} proposed the VGG-19 model which uses smaller convolutional filters and has depth of 19 layers. Since then the representational power and depth of these models have continued to grow every year. GoogleNet~\cite{GoogleNet} was a 22-layer network. In this paper, we perform all our experiments with the ResNet models proposed by He \etal~\cite{he2016resnet}. The core idea is to add residual connections between layers which helps in optimization of very-deep models. This results in new state-of-the-art performances on a number of recognition tasks.

Convolutional neural networks  learn a hierarchy of visual representations. These visual representations have been shown to be effective on a wide range of computer vision tasks~\cite{chen2016deeplab,FasterRCNN,RCNN,papandreou2017towards,DBLP:journals/corr/SimonyanZ14,DBLP:journals/corr/AgrawalGM14a,jain201515}. Learning these visual representations require large-scale training data. However, the biggest detection and segmentation datasets are still on the order of hundreds of thousands of images. Therefore, most of these approaches employ pre-training. The original model is learning using million labeled images in ImageNet and then further trained on  target tasks (fine-tuning) to yield better performance~\cite{RCNN, FasterRCNN, chen2016deeplab}. Huang \etal~\cite{Google_FRCNN} thoroughly evaluated the influence of multiple ConvNet architectures on object detection performance, and found that it is closely correlated with the models' capacity and classification performances on ImageNet.

While there has been significant work on increasing the representational capacity of ConvNets, the amount of training data for pre-training has remain kind of fixed over years. The prime reason behind this is the lack of human verified image datasets larger than ImageNet. In order to overcome the bottleneck, there have been recent efforts on visual representation learning using web-supervision~\cite{chen2015webly,chen2013neil,levan,DBLP:conf/mm/IzadiniaRFHH15,NIPS2010_0093,JoulinWeakly,DBLP:journals/corr/NiPBEBCW15,DBLP:journals/corr/KrauseSHZTDPL15} 
or unsupervised~\cite{Rubinstein13Unsupervised,doersch2015unsupervised,DBLP:journals/corr/DonahueKD16,DBLP:journals/corr/WangG15a,DBLP:journals/corr/PintoGHPG16,DBLP:journals/corr/PintoG15,DBLP:conf/nips/VondrickPT16} paradigms. However, most of these efforts are still are still exploratory in nature and far lower in performance compared to fully-supervised learning.

In this paper, we aim to shift the discussion from models to data. Our paper is inspired from several papers which have time and again paid closer look to impact and properties of data rather than models. In 2009, Pereira \etal~\cite{Pereira} presented a survey paper to look into impact of data in fields such as natural language processing and computer vision. They argued unlike physics, areas in AI are more likely to see an impact using more data-driven approaches. Another related work is the empirical study by Torralba and Efros~\cite{TorralbaEfros} that highlighted the dataset biases in current computer vision approaches and how it impacts future research.

Specifically, we focus on understanding the relationship between data and visual deep learning. There have been some efforts to understand this relationship. For example, Oquab \etal~\cite{OquabCVPR14} showed that expanding the training data to cover 1512 labels from ImageNet-14M further improves the object detection performance. Similarly, Huh \etal~\cite{HuhArXiv} showed that using a smaller subset of images for training from ImageNet hurts performance. Both these studies also show that selection of categories for training is important and random addition of categories tends to hurt the performance. But what happens when the number of categories are increased 10x? Do we still need manual selection of categories? Similarly, neither of these efforts demonstrated data effects at significantly larger scale.

Some recent work~\cite{JoulinWeakly,PlaNet} have looked at training ConvNets with significantly larger data. While~\cite{PlaNet} looked at geo-localization, \cite{JoulinWeakly} utilized the YFCC-100M dataset~\cite{YFCC100M} for representation learning. However, unlike ours, ~\cite{JoulinWeakly} showed plateauing of detection performance when trained on 100M images. Why is that? We believe there could be two possible reasons: a) YFCC-100M images come only from Flickr. JFT includes images all over the web, and has better visual diversity. The usage of user feedback signals in JFT further reduces label noise. YFCC-100M has a much bigger vocabulary size and noisier annotations. b) But more importantly, they did not see 
real effect of data due to use of smaller AlexNet of VGG models. In our experiments, we see more gain with larger model sizes.

\section{The JFT-300M Dataset}
We now introduce the JFT-300M dataset used throughout this paper. JFT-300M is a follow up version of the dataset introduced by~\cite{Hinton, Chollet}. The JFT-300M dataset is closely related and derived from the data which powers the Image Search. In this version, the dataset has 300M images and 375M labels, on average each image has 1.26 labels. These images are labeled with 18291 categories: \eg, {\bf 1165} type of animals and {\bf 5720} types of vehicles are labeled in the dataset. These categories form a rich hierarchy with the maximum depth of hierarchy being {\bf 12} and maximum number of child for parent node being {\bf 2876}.

\begin{figure}
\center
\includegraphics[width=\columnwidth]{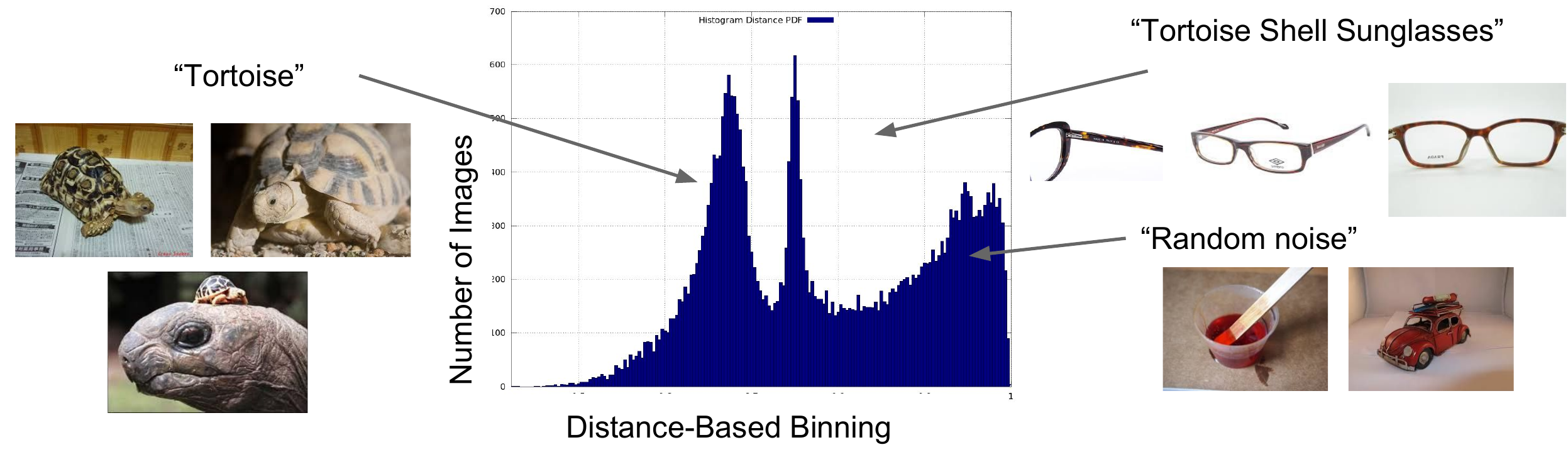}
\caption{JFT-300M dataset can be noisy in terms of label confusion and incorrect labels. This is because labels are generated via a complex mixture of web signals, and not annotated or cleaned by humans. x-axis corresponds to the quantized distances to K-Means centroids, which are computed based on visual features.}
\label{fig:noise}
\end{figure}

The images are labeled using an algorithm that uses complex mixture of raw web signals, connections between web-pages and user feedback. The algorithm starts from over one billion image label pairs, and ends up with 375M labels for 300M images with the aim to select labeled images with high precision. However, there is still some noise in the labels: approximately 20\% of the labels in this dataset are noisy. Since there is no exhaustive annotation, we have no way to estimate the recall of the labels. Figure~\ref{fig:noise} shows the kind of noise that exists in the dataset. Because the labels are generated automatically, there is a problem of `tortoise' being confused with `tortoise-shell glasses'.

Finally, it is important to discuss the data distribution of JFT-300M. The distribution is heavily long-tailed: \eg, there are more than {\bf 2M} `flowers', {\bf 3250} `subarau360' but only {\bf 131} images of `train conductors'. In fact, the tail is so heavy that we have more than 3K categories with less than 100 images each and approximately 2K categories with less than 20 images per category. 

\section{Training and Evaluation Framework}
We now describe our training and evaluation framework for the paper.

\subsection{Training on JFT-300M Data} \label{sec:jft_train_setup}
Although there are several novel ConvNet architectures recently proposed, we decide to use a standard Residual Network architecture~\cite{he2016resnet} with 101 layers (ResNet-101) for its state-of-the-art performance and the ease of comparison with previous work. To train a ResNet-101 model on JFT-300M, We add a fully-connected layer with 18291 outputs at the end of the network for classification. As the image labels are not mutually exclusive, we compute per-label logistic loss, and treat all non-present labels as negatives. To alleviate the issue of missing labels, we use a hand-designed label hierarchy and fill in the missing labels accordingly. For example, an image with label `apple' is also considered as a correct example for `fruit'.

During training, all input images are resized to $340\times 340$ pixels, and then randomly cropped to $299\times 299$. The image pixels are normalized to the range of $[-1, 1]$ independently per channel, and we use random reflection for data augmentation. We set weight decay to $10^{-4}$ and use batch normalization~\cite{BatchNorm} after all the convolutional layers. RMSProp optimizer is used with momentum of 0.9, and the batch size is set to 32. The learning rate is $10^{-3}$ initially and we decay it by 0.9 every 3M steps. We use asynchronous gradient descent training on 50 NVIDIA K80 GPUs. The model is implemented in TensorFlow.

To allow asynchrounous training of models on 50 GPUs, we adopt the Downpour SGD training scheme~\cite{DeanAsyncTraining}, where we use 17 parameter servers to store and update the model weights. 
The final classification fully-connected layer with 2048 input units and over 18K output units has over 36M parameters. To handle this in our parameter servers, we split it vertically into 50 equal sized sub-fc layers, and distribute them around different parameter servers.

\textbf{ImageNet baseline:} As observed by~\cite{Chollet}, hyperparameters that are selected to train with JFT-300M data yield sub-optimal performance when training on ImageNet (IVSVRC 2012 image classification dataset with 1.2M images). Therefore, for ImageNet, we use a momentum optimizer with the momentum of 0.9, and set the initial learning rate to $5\times 10^{-2}$ and batch size to 32. Learning rate is reduced by a factor of 10 every 30 epochs (~1.2M steps), and we train the model for a total of 5M steps. Similar to JFT-300M training, we use asynchronous gradient descent training on 50 NVIDIA K80 GPUs and 17 parameter servers. 

Our baseline ResNet-101 performs 1\% better than the open-sourced ResNet-101 checkpoint from the authors of~\cite{he2016resnet}, using the same evaluation protocol. 

\subsection{Monitoring Training Progress}
For monitoring the training progress on JFT-300M, we use the validation set from Chollet~\cite{Chollet}: `FastEval14k'. FastEval14k consists of 14000 images with labels from 6000 classes (subset of 18291 classes from JFT-300M). Unlike labels in JFT-300M, the images in FastEval14k are densely annotated and there are around 37 labels per image on average. We use the same mAP@100 metric as in~\cite{Chollet}, which is computed as the mean average precision (mAP) for top-100 predictions. Note that the class AP is weighted by how common the class is among social media images.


\begin{figure}[t]
\center
\includegraphics[width=0.6\columnwidth]{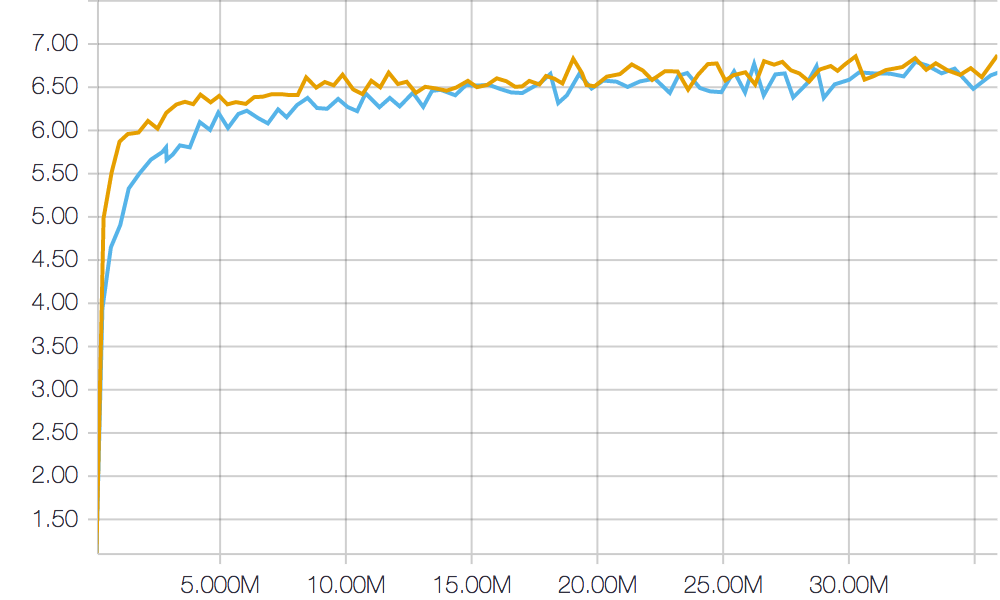}
\caption{Comparison of training progress with random initialization (blue) and ImageNet initialization (yellow) on JFT-300M data. x-axis is the number of training steps, and y-axis shows the mAP@100 metric computed on FastEval14k.}
\label{fig:300m_progress}
\end{figure}

We tried two strategies to initialize the model weights for training: random initialization and initializing from an ImageNet checkpoint. In both settings, we used the same training schedule (\eg, learning rates). We found that on FastEval14k benchmark, model trained from ImageNet initialization performs better at the first 15M iterations, but then becomes on par with random initialization.
Figure~\ref{fig:300m_progress} shows the training progress for these two settings. On FastEval14k benchmark, model trained from ImageNet initialization performs better at the first 15M iterations, but then becomes on par with random initialization.

Please note that the full training schedule takes 90M iterations or around 10 epochs. However, due to the time constraints, we train the models for 36M iterations or 4 epochs, which takes approximately 2 months. We will study the impact of training iterations in Section~\ref{sec:exp}.

\subsection{Evaluating the Visual Representations}
We use two approaches to evaluate the quality of visual representations learned from 300M training data. The first approach is to freeze the model weights and use these models as pure feature extractors. The second approach is to use the model weights as initialization and fine-tune the weights for other tasks. For evaluating visual representations, we select three representative computer vision tasks: object detection, semantic segmentation and human pose estimation.

We will perform a more rigorous ablative analysis to observe the effect of dataset size, vocabulary size, \etc on the object detection task. For the other tasks, we will just show how JFT-300M provides significant improvement compared to baseline ImageNet ResNet.\vspace{0.1in}

\noindent\textbf{De-duplication} One concern with using large-scale sets such as JFT-300M is the possible overlap between training and test sets. Such duplication exist in current frameworks as well: \eg 890 out of 50K validation images in ImageNet have  near-duplicate images training set. 
However, to ensure such duplication does not affect our results, we performed all experiments by removing near-duplicate images from test sets. We found the difference in performance to be insignificant for all the experiments. We therefore report de-duplicated test-set results in Appendix A.

\noindent\textbf{Object Detection.} We use the Faster RCNN framework~\cite{FasterRCNN} for its state-of-the-art performance. Faster RCNN is a two-stage model. The first stage is called region proposal network (RPN), which aims at generating class-agnostic object proposals. The second stage is a box classifier, it takes the boxes predicted by RPN and crops feature maps to generate classification predictions and refined bounding box predictions. These two stages share a common feature map generated by a ConvNet, and box classifier has additional convolutional layers before its final classification and regression layers. To use the ResNet-101 model pre-trained on JFT-300M data, we split the model into two parts: the first part starts from \textit{conv1} block and ends at \textit{conv4} block, it is used for feature extraction and is shared by both RPN and box classifier; the second part consists of the \textit{conv5} block, it is used by box classifier.\vspace{0.01in} 

\noindent\textbf{Semantic Segmentation.} We use the DeepLab framework~\cite{chen2016deeplab} with ResNet-101 base architecture for the task of semantic segmentation. In particular, we use a variant which adds four branches after the \textit{conv5} block of ResNet-101 architecture. Each branch is an atrous convolutional layer that predicts a sub-sampled pixel-wise class probabilities. Predictions from all branches are fused together to produce the final segmentation output. Please refer to the DeepLab-ASPP-L model (\textbf{A}trous \textbf{S}patial \textbf{P}yramid \textbf{P}ooling, with \textbf{L}arge atrous rates) from~\cite{chen2016deeplab} for details.

\noindent\textbf{Pose Estimation.}
We follow the framework proposed by Papandreou~\etal~\cite{papandreou2017towards}. It uses person bounding boxes detected by Faster RCNN, then applies a ResNet~\cite{he2016resnet} fully convolutionally to produce heatmaps and offsets for all keypoints. A novel scoring and non-maximum suppression (NMS) scheme is used to suppress duplicate detections and improve performance. We simply replace the base models used in their framework by our trained ResNet-101 models.

\section{Experiments}
\label{sec:exp}
We present results of fine-tuning JFT-300M ResNet-101 checkpoints on four tasks: image classification, object detection, semantic segmentation and human pose estimation.

\subsection{Image Classification}
We fine-tune the JFT-300M pre-trained ResNet101 using ImageNet classification data and compare it with a ResNet101 model trained from scratch. For this experiment, we use the standard ILSVRC 2012 `train' and `val' sets for training and evaluation. There are 1.2M training images and 50K validation images, over 1000 classes.

We use the same ImageNet training setup as described in Section~\ref{sec:jft_train_setup} for the ImageNet baseline,
but lowered the initial learning rate to $10^{-3}$ (standard for fine-tuning). We initialize the model weights from the JFT-300M checkpoint trained for 36M iterations and fine-tune on ImageNet for 4M iterations.

Table~\ref{tab:imagenet_ft} compares the fine-tuning results with models trained from the scratch. For reference, we show the random initialization performance for the open-sourced checkpoint from the authors of~\cite{he2016resnet}. We report top-1 and top-5 accuracies with a single crop being evaluated. We can see that fine-tuning
on JFT-300M gives considerable performance boost for both top-1 and top-5 accuracies. 

\begin{table}
\center
\begin{tabular}{l|r|r}
Initialization & Top-1 Acc. & Top-5 Acc. \\
\hline
MSRA checkpoint~\cite{he2016resnet} & 76.4 & 92.9 \\
Random initialization & 77.5 & 93.9 \\
Fine-tune from JFT-300M & \bf{79.2} & \bf{94.7} \\
\end{tabular}
\vspace{0.1in}
\caption{Top-1 and top-5 classification accuracy on the ImageNet `val' set (single model and single crop inference are used).}
\label{tab:imagenet_ft}
\end{table}

\subsection{Object Detection}
We next evaluate the JFT-300M checkpoints on object detection tasks. We evaluate on the two most popular datasets: COCO~\cite{COCO} and PASCAL VOC~\cite{Everingham10}. Instead of just showing state-of-the-art performance, we will also perform a rigorous ablative analysis to gain insights into the relationship between data and representation learning.

Specifically, we use object detection experiments to answer the following questions:
\begin{itemize}[noitemsep]
\item How does the performance of trained representations vary with iterations and epochs?
\item Does the performance of learned visual representations saturate after certain amount of data? Do we see any plateauing effect with more and more data?
\item How important is representational capacity?
\item Is the number of classes a key factor in learning visual representation?
\item How could clean data (\eg, ImageNet) help improve the visual representations?
\end{itemize} 

\subsubsection*{Experimental Setup}
For COCO~\cite{COCO}, we use a held-out 8000 images from the standard `val' set as our validation set, we refer to it as `\minival', the same set of images was used by~\cite{Google_FRCNN}. We use a combination of the standard training set and the remaining validation images for training. Unless otherwise specified, all COCO results are reported on the \minival\ set. In particular, we are interested in mean
average precision at 50\% IOU threshold (mAP@.5), and the average of mAP at IOU thresholds 50\% to 95\% (mAP@[.5, .95]). For our best ResNet101 models, we also evaluate on the COCO `test-dev' split (evaluated by the official result server). For PASCAL VOC, we use the 16551 `trainval' images from 
PASCAL VOC 2007 and 2012 for training, and report performance on the PASCAL VOC 2007 Test, which has 4952 images using mAP@.5 metric.


We use the TensorFlow Faster RCNN implementation~\cite{Google_FRCNN} and adopt their default training
hyperparameters except for learning rate schedules. We use asynchronous training with 9 GPU workers and 11 parameter servers, momentum optimizer is used with the momentum of 0.9. Each worker takes a single input image per step, the batch size for RPN and box classifier training
are 64 and 256 respectively. Input images are resized to have 600 minimum pixels and 1024 maximum pixels while maintaining the aspect ratio. The only data augmentation used is random flipping.

For COCO, we set the initial learning rate to be
$4\times 10^{-4}$, and decay the learning rate by a factor of 10 after 2.5M steps, the total number of steps is 3M.
For PASCAL VOC, we set the initial learning rate to be $3\times 10^{-4}$, and decay the learning rate by 0.1 after 500K steps, and the model is trained for 700K steps. The training schedules were selected on held-out validation images using the open-source ResNet-101 model (pre-trained on ImageNet). We found the same training schedules work well on other checkpoints, and keep them fixed throughout for fairer comparison. During inference, we use 300 RPN proposals. Our vanilla FasterRCNN implementation does not use the multi-scale inference, context or box-refinement as described in~\cite{FasterRCNN}.


\begin{table}
\begin{tabular}{l|c|c}
Method & mAP@0.5 & mAP@[0.5,0.95] \\
\hline
He \etal~\cite{he2016resnet} & 53.3 & 32.2 \\
ImageNet & 53.6 & 34.3 \\
300M & 56.9 & 36.7 \\
ImageNet+300M & \textbf{58.0} & \textbf{37.4} \\
\hline
Inception ResNet~\cite{InceptionResNet} & 56.3 & 35.5 \\
\end{tabular}
\vspace{0.01in}
\caption{Object detection performance comparisons with baseline methods on the COCO test-dev split. The first four Faster RCNN detectors are all based on ResNet-101 architecture, the last one is based on
the InceptionResNet-v2 architecture. During inference, a single image scale and crop, and a single detection model are used for all experiments. Vanilla Faster RCNN implementations are used for all systems except for He \etal~\cite{he2016resnet}, which also includes box refinement and context.} 
\label{tab:coco_test_dev}
\end{table}

\begin{table*}[t]
{\footnotesize
\renewcommand{\tabcolsep}{1.2mm}
\renewcommand{\arraystretch}{1.2}
\resizebox{\textwidth}{!}{
\begin{tabular}{@{} l | c | c | c | c | c | c | c | c | c | c | c | c | c | c | c | c | c | c | c | c | c@{}}
method & airplane & bicycle & bird & boat & bottle & bus & car & cat & chair & cow & table & dog & horse & mbike & person & plant & sheep & sofa & train & TV & mean\\
\hline
ImageNet & 79.7 & 80.6 & 77.1 & 65.9 & 64.2 & 85.3 & 81.0 & 88.4 & 60.5 & 83.1 & 70.8 & 86.7 & 86.2 & 79.7 & 79.5 & 49.5 & 78.3 & 80.2 & 79.2 & 69.7 & 76.3 \\
300M & 87.2 & 88.8 & 79.6 & 75.2 & 67.9 & 88.2 & 89.3 & 88.6 & 64.3 & 86.1 & 73.6 & 88.7 & 89.1 & 86.5 & 86.4 & 57.7 & 84.2 & 82.1 & 86.7 & 78.6 & 81.4\\
ImageNet+300M & 86.9 & 88.0 & 80.1 & 74.7 & 68.8 & 88.9 & 89.6 & 88.0 & 69.7 & 86.9 & 71.9 & 88.5 & 89.6 & 86.9 & 86.8 & 53.7 & 78.2 & 82.3 & 87.7 & 77.9 & 81.3 \\
\end{tabular}
}
}
\caption{Average Precision @ IOU threshold of 0.5 on PASCAL VOC 2007 `test' set. The `trainval' set of PASCAL VOC 2007 and 2012 are used for training.}
\label{tab:PASCAL_test}
\vspace{-0.1in}
\end{table*}

\subsubsection*{Comparison with ImageNet Models}
We first present the performance comparison with ImageNet checkpoints. Table~\ref{tab:coco_test_dev} shows the detection performance on COCO `test-dev' split. To show that our Faster RCNN baseline is competitive,
we also report results from the Faster RCNN paper~\cite{he2016resnet}, which uses both box refinement and context information. We can see that our ImageNet baseline performs competitively.

We evaluate JFT-300M trained from scratch (`300M') and from ImageNet initialization ('ImageNet+300M'). Both models outperforms the ImageNet baseline by large margins, with 3.3\% and 4.4\% boost in mAP@.5, 2.4\% and 3.1\% in mAP@[.5,.95] respectively. As a reference, we also show the performance of ImageNet trained InceptionResNetv2 in Table~\ref{tab:coco_test_dev}. We would like to point out that the gain is even more significant than recently achieved by doubling the number of layers on Inception ResNet~\cite{Google_FRCNN}. This clearly indicates that while there are indications of a plateauing effect on model representation capacity; in terms of data there is still a lot that can be easily gained.

Table~\ref{tab:PASCAL_test} shows the performance on the PASCAL VOC 2007 `test' set. Again, both JFT-300M checkpoints outperforms the ImageNet baseline significantly, by 5.1\% and 5.0\% mAP@.5 respectively.

\subsubsection*{Impact of Epochs}
We study how the number of training epochs affects the object detection performance. For this experiment we report results on COCO \minival\ set. Table~\ref{tab:jft_num_epoch} shows the performance comparison when the JFT-300M model has been trained for 1.3, 2.6 and 4 epochs respectively. We can see that as the number of training steps increases, the performance also improves. As a comparison, in Table~\ref{tab:in_num_epoch} we show the ImageNet counterpart when trained for 3, 6, 12 and 150 epochs, we can see that the
performance of ImageNet checkpoints improves faster than JFT-300M with respect to the number of epochs. 

We would also like to point out that our learning schedules have been developed using the experience from smaller datasets. One can envision better learning schedules which provide more improvement as more epochs are used.

\begin{table}
\center
\begin{tabular}{@{}l|r|r@{}}
\#Iters on JFT-300M & \#Epochs &  mAP@[0.5,0.95] \\
\hline
12M & 1.3 &  35.0  \\
24M & 2.6 & 36.1  \\
36M & 4 & 36.8  \\
\end{tabular}
\vspace{0.05in}
\caption{mAP@[.5,.95] on COCO \minival\ with JFT-300M checkpoint trained from scratch for different number of epochs.}
\label{tab:jft_num_epoch}
\end{table}

\begin{table}
\center
\begin{tabular}{@{}l|r|r@{}}
\#Iters on ImageNet & \#Epochs & mAP@[0.5,0.95] \\
\hline
100K & 3 & 22.2 \\
200K & 6 & 25.9 \\
400K & 12 & 27.4 \\
5M & 150 & 34.5 \\
\end{tabular}
\vspace{0.05in}
\caption{mAP@[.5,.95] on COCO \minival\ with ImageNet checkpoint trained for different number of epochs.}
\label{tab:in_num_epoch}
\vspace{-0.01in}
\end{table}

\subsubsection*{Impact of Data Size}\label{sec:impact_data}
For this experiment, we randomly sample a subset of 10M, 30M and 100M images from the JFT-300M training data. We use the same training schedule as the JFT-300M model training. We pick the checkpoints
corresponding to the 4th epoch for each subset. To study the impact of learned visual representations, we also conduct an experiments to freeze the model weights for all layers before the \textit{conv5} block. For this set of experiments we change the learning rate decay to happen at 900K steps, and the total number of training steps to 1.5M, as we find they tend to converge earlier.

\begin{figure}
\centering
\includegraphics[width=0.49\columnwidth]{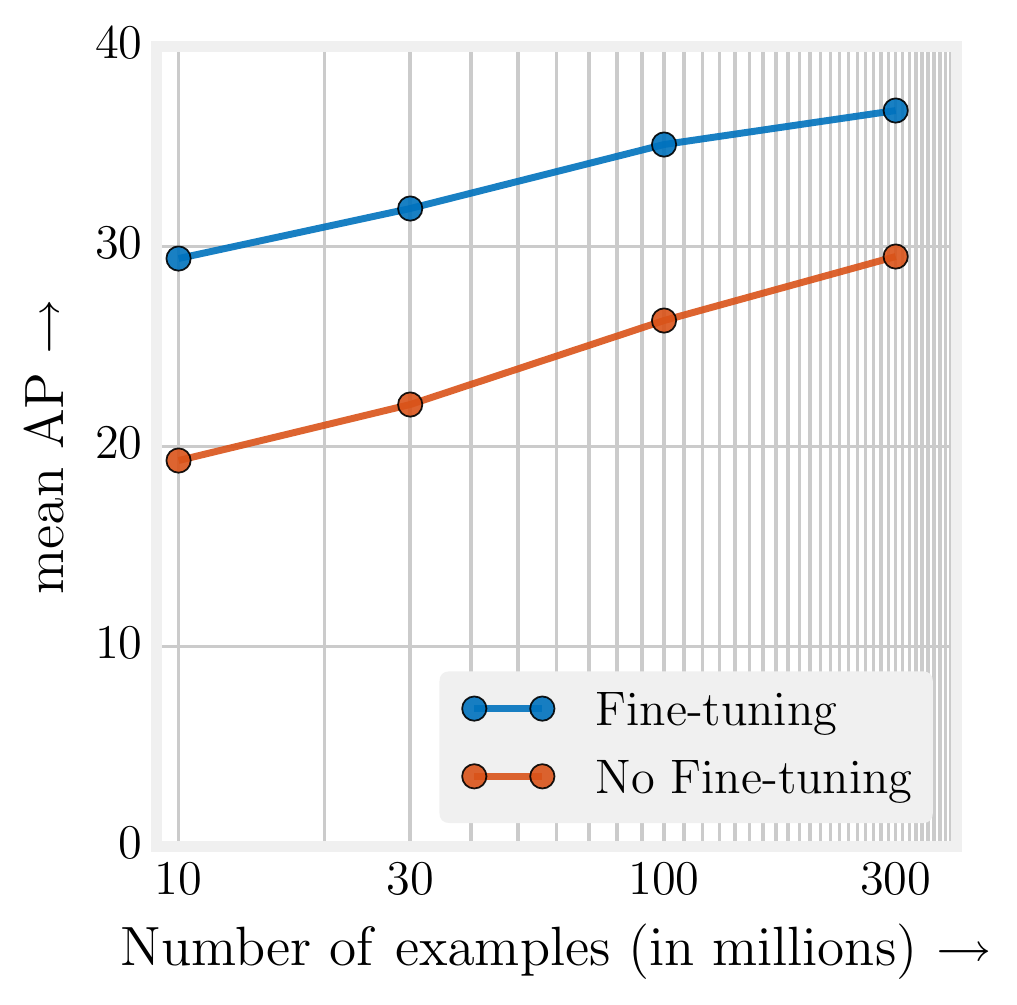}
\includegraphics[width=0.49\columnwidth]{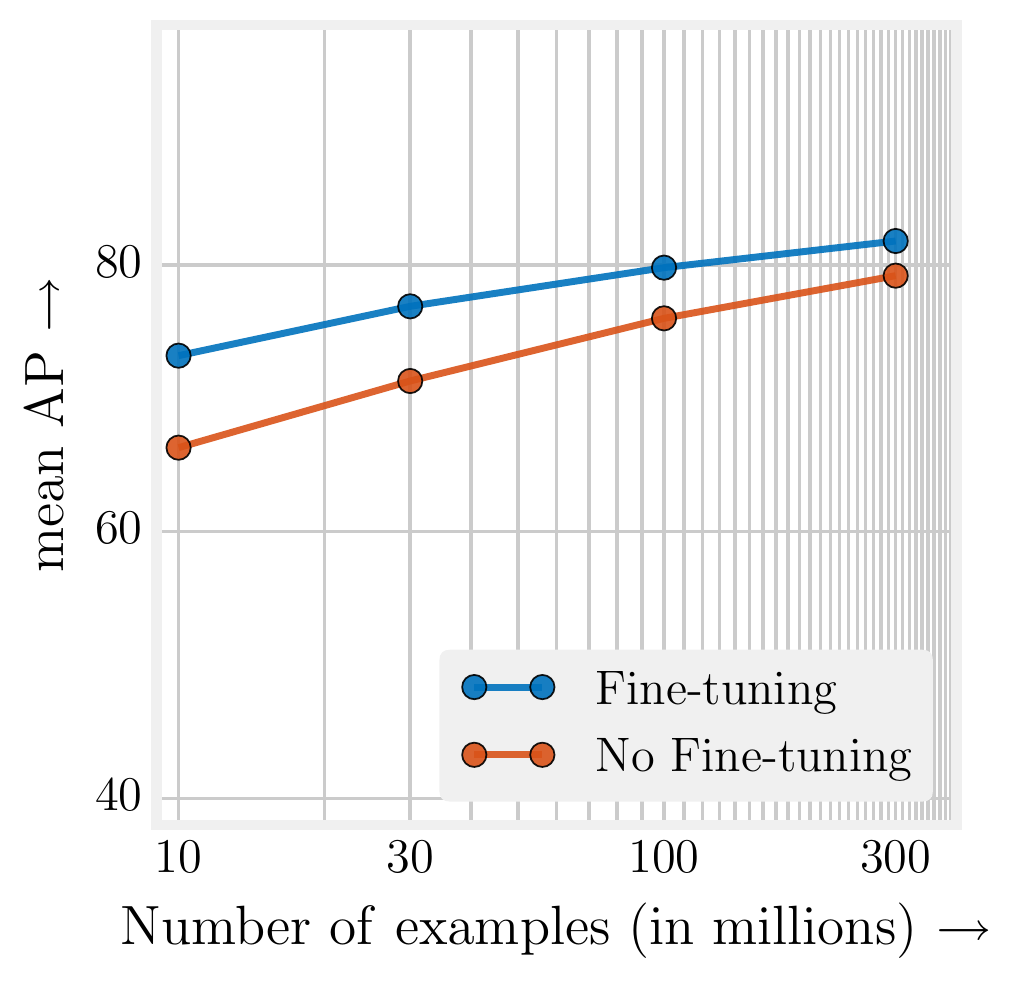}\\
\caption{Object detection performance when initial checkpoints are pre-trained on different subsets of JFT-300M from scratch. x-axis is the data size in log-scale, y-axis is the detection performance in mAP@[.5,.95] on COCO \minival\ (left), and in mAP@.5 on PASCAL VOC 2007 test (right).}
\label{fig:coco_data_size}
\end{figure}

In Figure~\ref{fig:coco_data_size}, we show the mAP@[.5,.95] with checkpoints trained on different JFT-300M subsets, the blue curve corresponds to the regular faster RCNN training (with fine-tuning), while the red curve corresponds to freezing feature extractors. Not surprisingly, fine-tuning offers significantly better performance on all data sizes. Most interestingly, we can see that the performance grows logarithmically as pre-training data expands, this is particularly true when feature extraction layers are frozen.

\subsubsection*{Impact of Classes}
JFT-300M has 18K labels in total. To understand what the large number of classes brings us, we select a subset of 941 labels which have direct correspondence to the 1000 ImageNet labels, and sample JFT-300M images which contain at least one of such labels. This results in a subset of 30M images. We then train on this dataset for 4 epochs using the same training scheme.

Table~\ref{tab:coco_num_classes} shows the performance comparison on COCO \minival\ set. We see that the two models perform on par with each other. This indicates that the performance benefit comes from more training images instead of more labels.

\begin{table}
\center
\begin{tabular}{@{}l|r@{}}
Number of classes     &  mAP@[.5,.95] \\
\hline
1K ImageNet & 31.2 \\
18K JFT & 31.9 \\
\end{tabular}
\vspace{0.05in}
\caption{Object detection performance in mean AP@[.5,.95] on COCO \minival\ set. We compare checkpoints pre-trained on 30M JFT images where labels are limited to the 1K ImageNet classes, and 30M JFT images covering all 18K JFT classes.}
\label{tab:coco_num_classes}
\end{table}

\subsubsection*{Impact of Model Capacity}

\begin{figure}
\centering
\begin{minipage}{0.25\textwidth}
   \centering
   \small
   \begin{tabular}{l | c | c}
\#Layers & ImageNet & 300M\\
\hline
50 & 31.6 & 33.5 \\
101 & 34.5 & 36.8 \\
152 & 34.7 & 37.7 \\
\end{tabular}
\end{minipage}%
\begin{minipage}{0.49\columnwidth}
   \centering
  \includegraphics[width=\columnwidth]{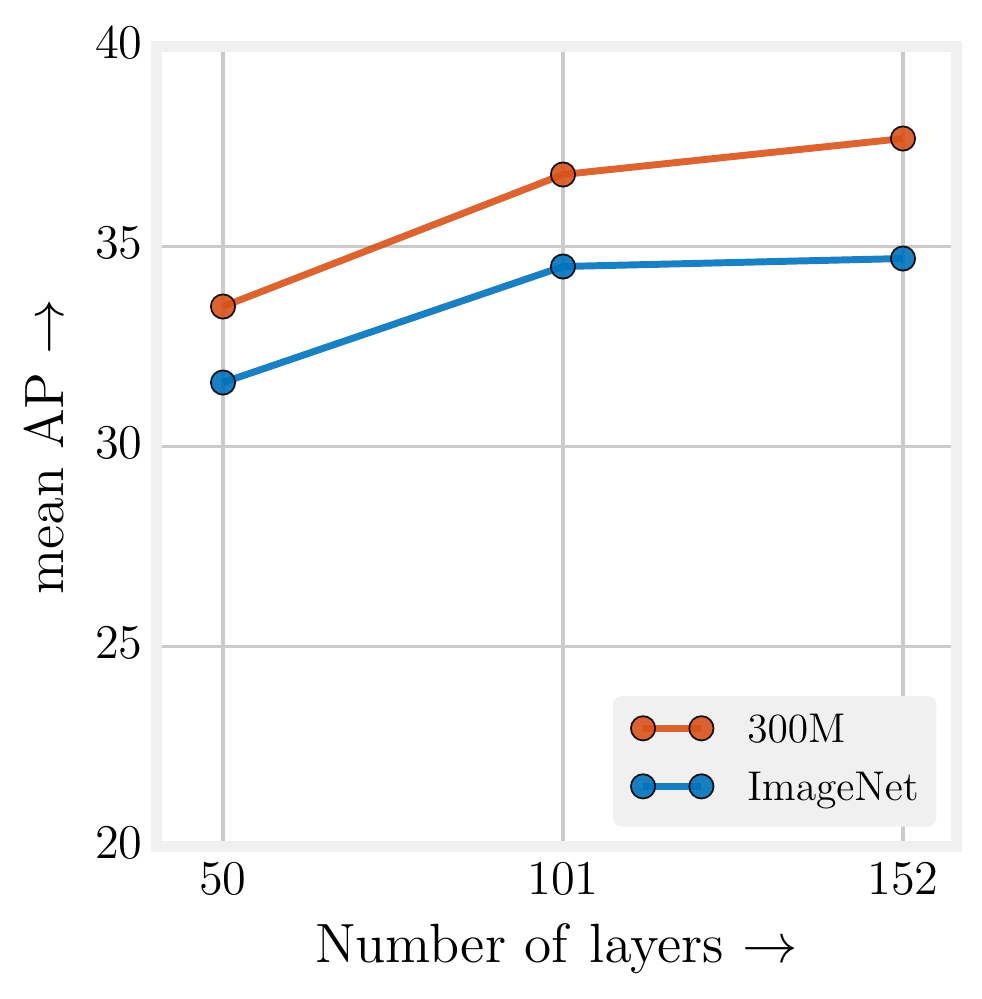}
\end{minipage}%
\vspace{0.01in}
\caption{Object detection performance on COCO \minival\ on ResNet models with different number of layers.}
\label{fig:coco_depth}
\end{figure}

Finally, we study the impact of model capacity when 300M images are available for training. We conduct the experiments on the 50-layer, 101-layer and 152-layer ResNet models. Each model is trained from scratch on the JFT-300M data, with the same hyper parameters used for ResNet-101 experiments. For comparison, we also train the models on ImageNet data till convergence, using the same hyper parameters for ResNet-101.

Figure~\ref{fig:coco_depth} shows the performance of fine-tuning different pre-trained models on COCO \minival set. We observe that higher capacity models are better at utilizing 300M data. For example, in case of ResNet-50 the gain is smaller compared to when using ResNet-152.

\subsection{Semantic Segmentation}
We use the PASCAL VOC 2012 semantic segmentation benchmark~\cite{voc} which has pixel-wise labels for 20 foreground classes and one background class. As is standard practice, all models are trained on an augmented PASCAL VOC~\cite{voc} 2012 `trainaug' set with 10582 images (extra annotations from~\cite{sbd}). We report quantitative results on the PASCAL VOC 2012 `val' set (1449 images) using the standard mean intersection-over-union (mIOU) metric.
 
\textbf{Implementation details}. The DeepLab-ASPP-L model~\cite{chen2016deeplab} has four parallel branches after \textit{conv5} block of ResNet101 architecture. Each branch is a $(3\times3)$ convolutional layer, with a different atrous rate $r$ ($r \in \{6, 12, 8, 24\}$). Different atrous rates enable the model to capture objects and context at different scales. Output of each branch is pixel-wise scores for 21 classes with the same resolution output map (subsampled by factor of 8 compared to the original image). These scores are added together and normalized for the final pixel-wise class probabilities.

For training, we use mini-batch SGD with momentum. Our model is trained for 30k SGD iterations using a mini-batch of 6 images, momentum of 0.9, an initialize learning rate (LR) of $10^{-3}$ and "polynomial" learning rate policy~\cite{chen2016deeplab}. All layers are trained with L2-regularization (weight decay of $5\times10^{-4}$). We do not use any data-augmentation, multi-scale training/testing or post-processing using CRFs for this task. To initialize the DeepLab-ASPP-L model using ImageNet or JFT-300M trained checkpoints, the final classification layer from these checkpoints is replaced with four convolutional branches (initialized using Xavier). All input images are resized to $(513\times513)$, which results in a $(65\times65)$ \textit{conv5} block from the ResNet101 network as well as $(65\times65\times21)$ predictions from the entire model.

\begin{figure}
\centering
\begin{minipage}{0.25\textwidth}
   \centering
   \begin{tabular}{@{}l | c@{}}
Initialization & mIOU\\
\hline
ImageNet & 73.6 \\
300M & 75.3 \\
ImageNet+300M & \textbf{76.5}\\
\end{tabular}
\end{minipage}%
\begin{minipage}{0.49\columnwidth}
   \centering
  \includegraphics[width=\columnwidth]{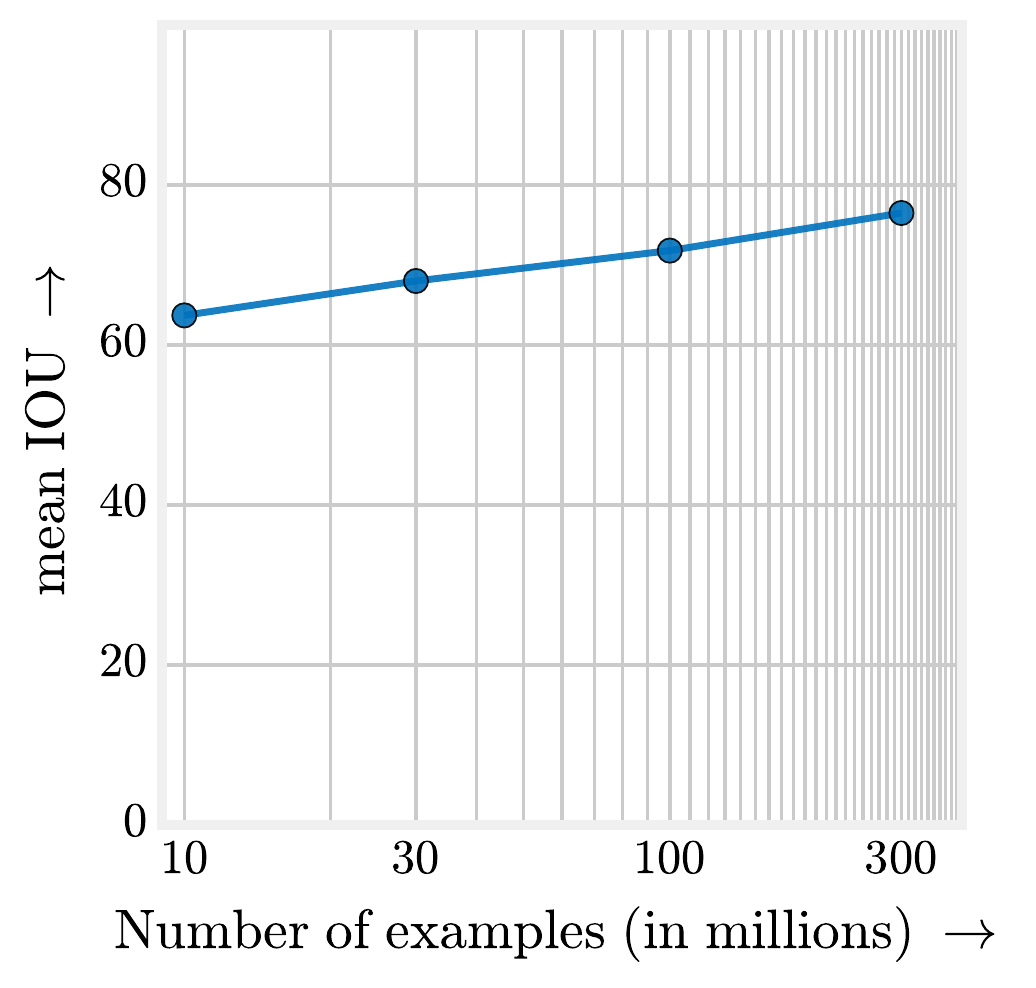}
\end{minipage}%
\vspace{0.01in}
\caption{Semantic segmentation performance on Pascal VOC 2012 val set. (left) Quantitative performance of different initializations; (right) Impact of data size on performance.}
\label{tab:segmentation}
\end{figure}

\paragraph{Comparison with ImageNet Models.} We present quantitative comparison of JFT-300M checkpoints with ImageNet checkpoints in Figure~\ref{tab:segmentation} (left). We see that the JFT-300M checkpoint outperforms ImageNet by 1.7\% points. We further observe that the JFT-300M model trained from the ImageNet checkpoint provides 2.9\% points boost over the vanilla ImageNet checkpoint.

\paragraph{Impact of Data Size.} In Figure~\ref{tab:segmentation} (right), we further present analysis of impact of training data size by randomly sampling a subset of 10M, 30M and 100M images from the JFT-300M for training base checkpoints (same as Section~\ref{sec:impact_data}). Once again we observe that the performance increases logarithmically as the pre-training dataset increases.

\begin{table}
\center
\begin{tabular}{@{} l | c | c | c | c}
& AP & AP@.5 & AR & AR@.5\\
\hline
CMU Pose~\cite{CMU_Pose} & 61.8 & 84.9 & 66.5 & 87.2 \\
\hline
ImageNet~\cite{papandreou2017towards} & 62.4 & 84.0 & 66.7 & 86.6\\
300M & \bf{64.8} & \bf{85.8} & \bf{69.4} & \bf{88.4} \\
ImageNet+300M & 64.4 & 85.7 & 69.1 & 88.2 \\
\end{tabular}
\vspace{0.1in}
\caption{Human pose estimation performance on COCO `test-dev' split. We follow the implementation of G-RMI Pose~\cite{papandreou2017towards}, but change the ResNet-101 initial checkpoints from ImageNet pre-trained to JFT-300M pre-trained.}
\label{tab:coco_pose}
\end{table}

\subsection{Human Pose Estimation}

We train the fully-convolutional pose detector~\cite{papandreou2017towards} by initializing the base ResNet model with our checkpoints and fine-tuning. The model is trained with SGD+Momentum for 450K steps. The learning rate was dropped by a factor of 10 after 250K steps, starting with a base learning rate. 
Best hyper parameter combination for each model was then selected independently and used in further experimentation.

In Table~\ref{tab:coco_pose}, we present the end to end pose estimation results evaluated on COCO `test-dev' set. G-RMI Pose uses the ImageNet pre-trained checkpoint for fine-tuning, and we can see that our models with JFT-300M initialization perform much better. Note that to have a fair comparison with G-RMI Pose, we show their performance when only COCO images are used for training (fine-tuning) and no ensembling is performed. We use the person detection results provided by the authors and apply our trained pose detectors on the same set of person boxes.

\section{Discussions}
Is it to be expected that performance of computer vision algorithms would always improve with more and more data? In our personal correspondences with several researchers, the general consensus seems to be that everyone expects some gain in performance numbers if the dataset size is increased dramatically, with decreasing marginal performance as the dataset grows. Yet, while a tremendous amount of time is spent on engineering and parameter sweeps; little to no time has been spent collectively on data. 

Our paper is an attempt to put the focus back on the data. The models seem to be plateauing but when it comes to the performance with respect to data -- but modest performance improvements are still possible for exponential increases of the data. Another major finding of our paper is that having better models is not leading to substantial gains because ImageNet is no more sufficient to use all the parameters or their representational power.

\vspace{-0.1in}
\paragraph{Representation learning:} One of the underlying debates is that should we spend more time collecting data for individual tasks such as detection and segmentation.
Our findings show there is still a lot to be gained from representation learning. Improved base models or base features can lead to significant gains in performance.

\vspace{-0.15in}
\paragraph{Disclaimer -- Large scale learning:} We would like to highlight that the training regime, learning schedules and parameters used in this paper are based on our understanding of training ConvNets with 1M images. Searching the right set of hyper-parameters requires significant more effort: even training a JFT model for 4 epochs needed 2 months on 50 K-80 GPUs. Therefore, in some sense the quantitative performance reported in this paper underestimates the impact of data for all reported image volumes.

\vspace{0.1in}
\small{\noindent {\bf Acknowledgements:}
This work would not have been possible without the heroic efforts of Image Understanding and Expander teams at Google who built the massive JFT dataset. We would specifically like to thank Tom Duerig, Neil Alldrin, Howard Zhou, Lu Chen, David Cai, Gal Chechik, Zheyun Feng, Xiangxin Zhu and Rahul Sukthankar for their help. Also big thanks to the VALE team for APIs and specifically, Jonathan Huang, George Papandreou, Liang-Chieh Chen and Kevin Murphy for helpful discussions.}


\small{
\bibliographystyle{ieee}
\bibliography{iccv2017.bib}
}

\newcommand{\vv}[1]{{\texttt{#1}}}
\newcommand{\conv}{\vv{conv}}
\newcommand{\fc}{\vv{fc}}
\newcolumntype{x}{>\small c}

\newcommand{\trainval}{trainval\raisebox{0.2ex}{$\ast$}}
\newcolumntype{L}[1]{>{\raggedright\let\newline\\\arraybackslash\hspace{0pt}}m{#1}}
\newcolumntype{C}[1]{>{\centering\let\newline\\\arraybackslash\hspace{0pt}}m{#1}}
\newcolumntype{R}[1]{>{\raggedleft\let\newline\\\arraybackslash\hspace{0pt}}m{#1}}
\newcommand{\hl}[1]{\underline{\textbf{#1}}}
\newcolumntype{o}{>\small L}

\clearpage
\onecolumn


\section*{Appendix A}

\subsubsection*{De-duplication Experiments}
A dataset with 300M images is almost guaranteed to contain images that overlap with the validation set of target tasks. In fact, we find that even for ImageNet, there are 890 out of 50K validation images have near-duplicate images in the training.

We use visual embeddings to measure similarities and identify duplicate or near-duplicate images. The embeddings are based on deep learning features. We find there are 5536 out of 50K images in ImageNet validation set, 1648 out of 8K images in COCO \minival, 201 out of 4952 images in Pascal VOC 2007 test set, and 84 out of 1449 images in Pascal VOC 2012 validation set that have near duplicates in JFT-300M. We rerun several experiments by removing near-duplicate images from validation sets and then comparing performance between baselines and learned models. We observe no significant differences in trends. Table~\ref{tab:imagenet_dedup},~\ref{tab:coco_dedup} and ~\ref{tab:pascal_dedup} show that the duplicate images have minimal impact on performance for all experiments.

\begin{table*}[h]
\centering
\renewcommand{\arraystretch}{1.1}
\renewcommand{\tabcolsep}{1.2mm}
\begin{tabular}{@{} l | c  c | c  c @{}}
& \multicolumn{2}{c|}{Original} & \multicolumn{2}{c}{De-duplication} \\
& Top-1 Acc. & Top-5 Acc. & Top-1 Acc. & Top-5 Acc. \\
\hline
MSRA checkpoint & 76.4 & 92.9 & 76.4 & 92.9 \\
Random initialization & 77.5 & 93.9 & 77.5 & 93.8 \\
Fine-tune from JFT-300M & 79.2 & 94.7 & 79.3 & 94.7 \\
\end{tabular}
\vspace{0.05in}
\caption{Top-1 and top-5 classification accuracy on ImageNet validation set, before and after de-duplication. Single model and single crop are used.}
\label{tab:imagenet_dedup}
\end{table*}
\vspace{-0.1in}

\begin{table*}[h]
\centering
\renewcommand{\arraystretch}{1.1}
\renewcommand{\tabcolsep}{1.2mm}
\begin{tabular}{@{} l | c  c | c  c @{}}
& \multicolumn{2}{c|}{Original} & \multicolumn{2}{c}{De-duplication} \\
& mAP@0.5 & mAP@[0.5,0.95] & mAP@0.5 & mAP@[0.5,0.95] \\
\hline
ImageNet & 54.0 & 34.5 & 54.0 & 34.6 \\
300M & 57.1 & 36.8 & 56.8 & 36.7 \\
ImageNet+300M & 58.2 & 37.8 & 58.2 & 37.7 \\
\end{tabular}
\vspace{0.07in}
\caption{mAP@0.5 and mAP@[0.5,0.95] for object detection performance on COCO \minival, before and after de-duplication.}
\label{tab:coco_dedup}
\end{table*}
\vspace{-0.1in}

\begin{table*}[h]
\centering
\renewcommand{\arraystretch}{1.1}
\renewcommand{\tabcolsep}{1.2mm}
\begin{tabular}{@{} l | c  c | c  c @{}}
& \multicolumn{2}{c|}{VOC07 Detection} & \multicolumn{2}{c}{VOC12 Segmentation} \\
& Original & De-duplication & Original & De-duplication \\
\hline
ImageNet & 76.3 & 76.5 & 73.6 & 73.3 \\
300M & 81.4 & 81.5 & 75.3 & 75.1 \\
ImageNet+300M & 81.3 & 81.2 & 76.5 & 76.5 \\
\end{tabular}
\vspace{0.05in}
\caption{Object detection and semantic segmentation performance on Pascal VOC, before and after deduplication. (Left) Object detection mAP@0.5 on Pascal VOC 2007 test set. (Right) Semantic segmentation mIOU on Pascal VOC 2012 validation set.}
\label{tab:pascal_dedup}
\end{table*}

We do not conduct de-duplication experiments of COCO testdev dataset for object detection and pose estimation as their groundtruth annotations are not publicly available.

\clearpage

\section*{Appendix B}

\subsubsection*{Detailed and Per-category Results: Object Detection}
In this section, we present detailed and per-category object detection results for Table 2 (Section 5.2) from the main submission, evaluated on the COCO test-dev split. In Table~\ref{tab:coco_main}, we report detailed AP and AR results using different initializations. In Table~\ref{tab:per_category}, we provide per-category AP and AP@.5 results.

\begin{table*}[h]
\centering
\renewcommand{\arraystretch}{1.1}
\renewcommand{\tabcolsep}{1.2mm}
\begin{tabular}{@{} l | c  c  c  c  c  c | c  c  c  c  c  c @{}}

& AP & AP@.5 & AP@.75 & AP(S) & AP(M) & AP(L) & AR & AR@.5 & AR@.75 & AR(S) & AR(M) & AR(L)\\
\hline
ImageNet & 34.3 & 53.6 & 36.9 & 15.1 & 37.4 & 48.5 & 30.2 & 47.3 & 49.7 & 26.0 & 54.6 & 68.6\\
300M & 36.7 & 56.9 & 39.5 & 17.1 & 40.0 & 50.7 & 31.5 & 49.3 & 51.9 & 28.6 & 56.9 & 70.4 \\
ImageNet+300M & 37.4 & 58.0 & 40.1 & 17.5 & 41.1 & 51.2 & 31.8 & 49.8 & 52.4 & 29.0 & 57.7 & 70.5 \\
\end{tabular}
\vspace{0.1in}
\caption{Object detection performance on COCO test-dev split using different model initializations.}
\label{tab:coco_main}
\end{table*}
\vspace{-0.1in}

\subsubsection*{Per-category Results: Semantic Segmentation}
In Table~\ref{tab:segmentation_supp}, we report quantitative results on the VOC 2012 segmentation validation set for all classes (refer to Figure 5 (left), Section 5.3 in the main submission). Results are reported for different initializations. We observe more than 7 point improvement for categories like boat and horse.

\begin{table*}[h]
	\centering
	\renewcommand{\arraystretch}{1.1}
	\renewcommand{\tabcolsep}{1.2mm}
	\resizebox{\linewidth}{!}{
		\begin{tabular}{@{}L{2.5cm} c c*{29}{x} @{}}
			\toprule
			Initialization & mIOU & bg & {aero} & {bike} & {bird} & {boat} & {bottle} & {bus} & {car}& {cat} & {chair} & {cow} & {table} & {dog} & {horse} & {mbike} & {persn} & {plant} & {sheep} & {sofa} & {train} & {tv} \\
			\midrule
			ImageNet &
			73.6 & 93.2 & 88.9 & 40.1 & 87.3 & 65.0 & 78.8 & 89.9 & 84.3 & 88.8 & 37.2 & 81.6 & 49.3 & 84.1 & 78.9 & 79.3 & 83.3 & 57.7 & 82.0 & 41.7 & 80.3 & 73.1 \\
			300M &
			75.3 & 93.7 & 89.8 & 40.1 & 89.8 & 70.6 & 78.5 & 89.9 & 86.1 & 92.0 & 36.9 & 80.9 & 52.8 & 87.6 & 82.4 & 80.8 & 84.3 & 61.7 & 84.4 & 44.8 & 80.9 & 72.6\\
			ImageNet+300M &
			76.5 & 94.8 & 90.4 & 41.6 & 89.1 & 73.1 & 80.4 & 92.3 & 86.7 & 92.0 & 39.6 & 82.7 & 52.7 & 86.2 & 86.1 & 83.6 & 85.7 & 61.5 & 83.9 & 45.3 & 84.6 & 73.6 \\
			\bottomrule
		\end{tabular}
	}
	\vspace{0.01in}
	\caption{Per-class semantic segmentation performance on PASCAL VOC 2012 validation set.}
	\label{tab:segmentation_supp}
\end{table*}
\vspace{-0.1in}

\subsubsection*{Detailed Results: Human Pose Estimation}
In Table~\ref{tab:coco_pose_supp}, we present all AP and AR results for the performance reported in Table 7 (Section 5.4) in the main submission.

\begin{table*}[h]
\centering
\begin{tabular}{@{} l | c  c  c  c  c | c  c  c  c  c @{}}
& AP & AP@.5 & AP@.75 & AP(M) & AP(L) & AR & AR@.5 & AR@.75 & AR(M) & AR(L)\\
\hline
CMU Pose~[3] & 61.8 & 84.9 & 67.5 & 57.1 & 68.2 & 66.5 & 87.2 & 71.8 & 60.6 & 74.6 \\
ImageNet~[26] & 62.4 & 84.0 & 68.5 & 59.1 & 68.1 & 66.7 & 86.6 & 72.0 & 60.8 & 74.9\\
300M & 64.8 & 85.8 & 71.5 & 62.2 & 70.3 & 69.4 & 88.4 & 75.2 & 63.9 & 77.0\\
ImageNet+300M & 64.4 & 85.7 & 70.7 & 61.8 & 69.8 & 69.1 & 88.2 & 74.8 & 63.7 & 76.6\\
\end{tabular}
\vspace{0.05in}
\caption{Human pose estimation performance on the COCO test-dev split.}
\label{tab:coco_pose_supp}
\end{table*}

\begin{table*}[h!]
\centering
\caption{Per-class object detection performance on COCO test-dev split using different model initializations.}
\label{tab:allclasses}
\vspace{0.1in}
\renewcommand{\arraystretch}{1.2}
\renewcommand{\tabcolsep}{1.2mm}
\resizebox{0.485\linewidth}{!}{
\footnotesize
\begin{tabular}{@{}
L{1.7cm} 
!{\color{gray}\vrule} cc
!{\color{gray}\vrule} cc
!{\color{gray}\vrule} cc
@{}}
\Xhline{1pt}
\multicolumn{1}{r!{\color{gray}\vrule}}{Initialization $\rightarrow$} & \multicolumn{2}{c!{\color{gray}\vrule}}{ImageNet} & \multicolumn{2}{c!{\color{gray}\vrule}}{300M} &
\multicolumn{2}{c}{{\footnotesize ImageNet+300M}}\\
& \scriptsize AP@.5 & \scriptsize AP &  \scriptsize AP@.5 & \scriptsize AP & \scriptsize AP@.5 & \scriptsize AP \\ 
\Xhline{0.5pt}
person & 71.5 & 47.7 & 73.1 & 49.8 & 72.7 & 49.9 \\
bicycle & 48.9 & 26.4 & 54.9 & 30.0 & 52.7 & 29.9 \\
car & 55.7 & 34.7 & 58.3 & 36.9 & 59.3 & 37.1 \\
motorcycle & 56.5 & 36.7 & 61.6 & 40.5 & 59.9 & 39.6 \\
airplane & 67.9 & 52.0 & 70.1 & 55.0 & 70.4 & 54.7 \\
bus & 77.7 & 62.5 & 79.5 & 64.6 & 79.0 & 64.2 \\
train & 66.8 & 59.2 & 69.7 & 62.8 & 69.7 & 62.1 \\
truck & 46.3 & 29.9 & 49.7 & 33.0 & 52.2 & 34.5 \\
boat & 30.6 & 19.4 & 32.5 & 22.1 & 32.1 & 22.3 \\
traffic light & 48.9 & 22.7 & 49.8 & 24.3 & 49.1 & 24.6 \\
fire hydrant & 75.3 & 59.1 & 74.4 & 59.3 & 74.9 & 59.5 \\
stop sign & 83.2 & 63.6 & 84.4 & 63.8 & 85.6 & 66.4 \\
parking meter & 62.2 & 37.5 & 64.9 & 38.5 & 64.5 & 37.6 \\
bench & 38.1 & 19.6 & 39.3 & 20.1 & 40.6 & 21.4 \\
bird & 60.2 & 29.4 & 61.9 & 33.0 & 63.3 & 34.2 \\
cat & 64.2 & 58.1 & 68.0 & 61.9 & 67.9 & 62.4 \\
dog & 62.6 & 52.9 & 66.1 & 56.2 & 66.9 & 57.3 \\
horse & 67.2 & 53.5 & 70.8 & 57.0 & 71.3 & 57.0 \\
sheep & 64.4 & 43.6 & 64.8 & 45.4 & 66.7 & 46.3 \\
cow & 70.7 & 45.4 & 71.9 & 47.4 & 73.3 & 48.9 \\
elephant & 75.1 & 64.1 & 77.3 & 66.4 & 76.1 & 65.5 \\
bear & 70.5 & 66.9 & 74.5 & 69.8 & 72.7 & 70.0 \\
zebra & 71.0 & 59.3 & 71.5 & 60.4 & 71.3 & 61.0 \\
giraffe & 75.3 & 67.4 & 75.9 & 69.0 & 75.9 & 69.3 \\
backpack & 19.6 & 12.8 & 19.5 & 14.7 & 18.5 & 15.1 \\
umbrella & 46.2 & 28.9 & 50.7 & 32.3 & 50.4 & 32.8 \\
handbag & 14.7 & 9.7 & 13.7 & 10.9 & 16.1 & 12.0 \\
tie & 50.8 & 26.3 & 53.2 & 27.9 & 51.5 & 28.4 \\
suitcase & 40.4 & 26.7 & 44.4 & 30.3 & 46.9 & 32.5 \\
frisbee & 53.4 & 43.8 & 55.3 & 48.6 & 58.6 & 48.3 \\
skis & 1.5 & 18.1 & 3.0 & 20.0 & 2.3 & 20.7 \\
snowboard & 45.7 & 29.3 & 47.0 & 33.3 & 43.9 & 32.1 \\
sports ball & 41.8 & 35.6 & 48.7 & 37.6 & 42.3 & 38.6 \\
kite & 39.4 & 37.5 & 33.9 & 38.9 & 35.9 & 40.0 \\
baseball bat & 8.3 & 23.4 & 6.7 & 25.1 & 9.9 & 27.5 \\
baseball glove & 35.6 & 27.4 & 33.7 & 31.2 & 41.9 & 31.8 \\
skateboard & 42.2 & 40.0 & 48.6 & 44.7 & 49.2 & 44.4 \\
surfboard & 48.5 & 31.1 & 51.7 & 32.8 & 52.4 & 33.9 \\
tennis racket & 53.1 & 42.6 & 55.1 & 44.1 & 55.4 & 45.1 \\
bottle & 61.2 & 28.6 & 61.8 & 30.5 & 61.6 & 30.8 \\
\Xhline{1pt}
\end{tabular}
}
\hspace{0.2cm}
\resizebox{0.485\linewidth}{!}{
\footnotesize
\begin{tabular}{@{}
L{1.7cm} 
!{\color{gray}\vrule} cc
!{\color{gray}\vrule} cc
!{\color{gray}\vrule} cc
@{}}
\Xhline{1pt}
\multicolumn{1}{r!{\color{gray}\vrule}}{Initialization $\rightarrow$} & \multicolumn{2}{c!{\color{gray}\vrule}}{ImageNet} & \multicolumn{2}{c!{\color{gray}\vrule}}{300M} &
\multicolumn{2}{c}{{\footnotesize ImageNet+300M}}\\
& \scriptsize AP@.5 & \scriptsize AP &  \scriptsize AP@.5 & \scriptsize AP & \scriptsize AP@.5 & \scriptsize AP \\ 
\Xhline{0.5pt}
wine glass & 53.8 & 30.2 & 56.3 & 33.3 & 58.7 & 34.7 \\
cup & 64.7 & 32.5 & 67.5 & 35.6 & 68.4 & 35.9 \\
fork & 45.7 & 23.2 & 45.1 & 26.5 & 50.1 & 27.8 \\
knife & 29.9 & 12.8 & 37.1 & 15.7 & 37.2 & 16.4 \\
spoon & 13.0 & 10.0 & 11.4 & 11.7 & 11.6 & 13.3 \\
bowl & 49.4 & 32.1 & 53.6 & 35.4 & 52.2 & 35.4 \\
banana & 38.1 & 18.7 & 39.8 & 20.4 & 40.0 & 21.1 \\
apple & 49.4 & 19.1 & 50.1 & 20.8 & 51.5 & 21.7 \\
sandwich & 44.0 & 29.6 & 45.2 & 31.3 & 47.8 & 34.1 \\
orange & 48.7 & 25.0 & 50.7 & 26.2 & 49.0 & 26.1 \\
broccoli & 30.6 & 22.9 & 32.5 & 24.8 & 31.9 & 24.6 \\
carrot & 25.9 & 14.0 & 28.6 & 16.1 & 21.5 & 16.4 \\
hot dog & 43.7 & 21.8 & 46.5 & 24.8 & 48.2 & 25.8 \\
pizza & 67.9 & 51.1 & 69.0 & 52.3 & 68.7 & 52.8 \\
donut & 60.2 & 40.1 & 64.8 & 43.9 & 66.8 & 46.4 \\
cake & 42.7 & 25.5 & 46.4 & 28.1 & 46.5 & 29.1 \\
chair & 33.0 & 21.1 & 36.7 & 24.0 & 35.9 & 24.4 \\
couch & 41.3 & 36.2 & 44.5 & 38.9 & 44.9 & 39.4 \\
potted plant & 25.6 & 20.1 & 27.3 & 21.9 & 30.0 & 23.4 \\
bed & 44.5 & 40.6 & 45.6 & 41.7 & 47.2 & 43.4 \\
dining table & 33.9 & 25.3 & 36.3 & 27.5 & 36.8 & 27.6 \\
toilet & 61.1 & 54.8 & 61.8 & 56.1 & 63.3 & 57.4 \\
tv & 61.8 & 50.0 & 63.0 & 51.9 & 63.7 & 52.7 \\
laptop & 65.8 & 54.5 & 68.3 & 56.6 & 68.9 & 57.5 \\
mouse & 72.1 & 44.4 & 72.0 & 47.6 & 75.6 & 47.3 \\
remote & 56.4 & 22.1 & 55.8 & 24.4 & 59.1 & 26.0 \\
keyboard & 57.1 & 45.4 & 57.5 & 45.9 & 61.4 & 48.3 \\
cell phone & 54.0 & 23.4 & 58.5 & 26.1 & 57.5 & 26.7 \\
microwave & 53.9 & 50.3 & 53.7 & 50.5 & 58.7 & 53.1 \\
oven & 40.9 & 31.7 & 41.9 & 33.5 & 43.2 & 34.6 \\
toaster & 32.6 & 14.7 & 39.9 & 20.5 & 32.9 & 20.1 \\
sink & 43.2 & 31.0 & 44.8 & 34.4 & 44.0 & 33.9 \\
refrigerator & 48.6 & 42.3 & 51.7 & 44.6 & 52.4 & 46.1 \\
book & 15.2 & 7.4 & 18.7 & 8.8 & 21.3 & 9.8 \\
clock & 56.7 & 43.7 & 56.7 & 45.3 & 55.8 & 45.1 \\
vase & 57.1 & 32.3 & 61.5 & 35.9 & 61.4 & 36.5 \\
scissors & 31.1 & 20.8 & 38.9 & 25.2 & 34.8 & 25.9 \\
teddy bear & 50.4 & 35.4 & 54.7 & 40.2 & 54.7 & 40.4 \\
hair drier & 2.3 & 1.0 & 4.8 & 1.8 & 4.0 & 1.9 \\
toothbrush & 48.5 & 34.3 & 50.7 & 36.7 & 51.2 & 37.4 \\
\Xhline{1pt}
\end{tabular}
}
\label{tab:per_category}
\end{table*}

\end{document}